%% file: ijca2017.tex
\documentclass[11pt,a4paper]{article}
\usepackage[hyperref]{acl2018}

\aclfinalcopy
\usepackage{times}
\usepackage{latexsym}
\usepackage{caption}
 \usepackage{amsmath}
 \usepackage{amsfonts}
 \usepackage{algorithmicx}
\usepackage{subfig}
\usepackage{xcolor}

\usepackage{tikz}
\usepackage{pgfplots}
\pgfplotsset{compat=1.7}
\usepackage{pgfplotstable}
\usepackage{balance}
\pgfplotstableread{dat/baseline.txt}\performanceBaseline
\usepgfplotslibrary{fillbetween}
\usepackage{algorithm}
\pgfplotscreateplotcyclelist{my black white}{%
	solid ,black, every mark/.append style={solid, fill=white, black}, mark=o\\%
	thick, solid, black, every mark/.append style={solid,mark size=3,red, fill=white}, mark=star\\%
	solid, black, every mark/.append style={solid, fill=black!60!brown}, mark=*\\%
	solid, black, every mark/.append style={solid, fill=gray, black, mark size=3}, mark=diamond*\\%
	dashed, every mark/.append style={solid, fill=gray},mark=triangle*\\%
	loosely dashed, every mark/.append style={solid, fill=gray},mark=*\\%
	densely dashed, every mark/.append style={solid, fill=gray},mark=square*\\%
	dashdotted, every mark/.append style={solid, fill=gray},mark=otimes*\\%
	dasdotdotted, every mark/.append style={solid},mark=star\\%
	densely dashdotted,every mark/.append style={solid, fill=gray},mark=diamond*\\%
}
\usepackage{listings} 
\usetikzlibrary{plotmarks}
\usepackage{etoolbox}
\usetikzlibrary{patterns}
\usepackage{multirow}
\usepgfplotslibrary{groupplots}
\usepackage{algorithm} 
\usepackage[noend]{algpseudocode}
\usepackage{algorithmicx}
\algnewcommand\algorithmicforeach{\textbf{for each}}
\algdef{S}[FOR]{ForEach}[1]{\algorithmicforeach\ #1\ \algorithmicdo}
\algnewcommand\algorithmicoptional{\textbf{Optional:}}
\algnewcommand\OPTIONAL{\item[\algorithmicoptional]}
\makeatletter
\def\BState{\State\hskip-\ALG@thistlm}
\makeatother

\let\oldbibitem\bibitem
\def\bibitem{\vfill\oldbibitem}

\title{Sketching Word Vectors Through Hashing}

\author{
Behrang QasemiZadeh \\
DFG SFB 991 \\
Universit\"{a}t D\"{u}sseldorf \\
D\"{u}sseldorf, Germany \\
{\small \tt zadeh@phil.hhu.de} \\
\And
Laura Kallmeyer \\
DFG SFB 991 \\
Universit\"{a}t D\"{u}sseldorf \\
D\"{u}sseldorf, Germany \\
{\small \tt kallmeyer@phil.hhu.de}} 

\makeatletter

\patchcmd{\NAT@test}{\else \NAT@nm}{\else \NAT@nmfmt{\NAT@nm}}{}{}

\DeclareRobustCommand\citepos
  {\begingroup
   \let\NAT@nmfmt\NAT@posfmt
   \NAT@swafalse\let\NAT@ctype\z@\NAT@partrue
   \@ifstar{\NAT@fulltrue\NAT@citetp}{\NAT@fullfalse\NAT@citetp}}

\let\NAT@orig@nmfmt\NAT@nmfmt
\def\NAT@posfmt#1{\NAT@orig@nmfmt{#1's}}

\makeatother

\begin{document}
\maketitle
\begin{abstract}
We present a new method for building word vectors using a derandomization of a novel random projection. We exploit the fact that counting co-occurrences of words with a set of randomly generated bag-of-words yields linguistically informative distributional representations of words. We generate these bag-of-words using the modulus of the hash codes assigned to the words. This boils down to an elegant and flexible algorithm for learning word vectors. We show that these word vectors achieve results competitive to the ones produced using neural networks in both upstream and downstream NLP tasks, but just faster, easier, and with a considerably less amount of computations.


\end{abstract}

\section{Introduction}

Random patterns are helpful for learning word representations. Methods for learning word representations (i.e., using distributional frequencies to produce word vectors of reduced dimension, e.g., \newcite{Mikolov2013,glove2014,bojanowski2017enriching}, to name just a few), a cornerstone of modern data-driven NLP, are usually motivated by \citepos{Harris1954distributional} Distributional Hypothesis (DH) that words of comparable linguistic properties appear with/within a similar set of `contexts/structures'; for instance, words of similar meaning co-occur with a similar set of context words $C=\{c_1, \ldots c_n\}$. This narration of the hypothesis implies that if $C$ is partitioned randomly into $m$ buckets, e.g. $\{\{c_1\ldots c_x\}_1, \ldots, \{c_y, \ldots c_n\}_m\}$, co-related words co-occur with similar buckets, too. We exploit this reading of DH to propose our method for word representation learning. 

In our method, context words are assigned to buckets that are generated randomly. The random generation of buckets is replicated using hash functions. Instead of counting co-occurrences of a word with other context words in a corpus, we keep track of the count of the co-occurrences of words and the buckets that context words are assigned to. We provide empirical evidence that the distributional patterns in these word-per-random-bucket counts can be used, e.g., to perform reasonably well in a range of so-called word relatedness tests, and to provide sufficient information about words for neural networks that solve NLP problems, i.e. to achieve performance competitive to successful neural word representation learning methods but without solving a complex optimization problem.


In short, the main contribution of this paper is a word representation learning algorithm which is astonishingly simple and has low computational complexity but it yet shows performance competitive to sophisticated methods of word representation learning. In the reminder of this paper, in Section~\ref{sec:method}, we propose our new hash-based method for word representation learning, in which we describe our method as a dimension reduction process using random projections (i.e., a matrix multiplication computed using the distributive property of multiplication over addition), and further explain the novelty of our method concerning its underlying random projections. Section~\ref{relatedwork} describes related work. Section~\ref{sec:evaluation} reports results from a number of experiments, in which word vectors are used directly or indirectly to deliver results. Section~\ref{conclusion} concludes this paper.

\section{The Proposed Hash-based Method}\label{sec:method}
Lets assume that we want to build an $m$-dimensional vector representation for a word $w$ that is co-occurred with a large number $n$ ($m \ll n$) of context words $c$. To build a vector for each $w$ (denoted by $\vec{w}$), our method takes the following steps: 
\begin{algorithm}\captionsetup{labelfont={sc,bf}}
  \caption{Building Word Vectors}\label{algorithm1}
\begin{algorithmic}[1]
\State Initialize a zero $m$-dimensional vector $\vec{w}$
\ForEach {$c$ co-occurred with $w$}
	\State \texttt{d} $\gets$ \texttt{abs}(\texttt{hash}($c$) \texttt{\%} $m$)
	\State $\vec{w}_{\texttt{d}} = \vec{w}_{\texttt{d}} + 1$
\EndFor
\Return $\vec{w}$
\end{algorithmic}
\end{algorithm}\\
Above, $w_{\texttt{d}}$ is the $d$th component of $\vec{w}$. The \texttt{hash} function assigns hash codes (e.g., an integer) to each context word; ideally the generated hash codes are independently and identically distributed. The \texttt{abs} function returns the absolute value of its input number; \texttt{\%} is the modulus operator and it gives the remainder of the division of the generated hash code by the chosen value $m$. In our implementation, we choose the following hash function since it shows a low collision rate for short words (bytes sequences):\footnote{See \url{http://www.burtleburtle.net/bob/hash/doobs.html}, the Jenkins hash functions.}
\lstset{language=C++,
                basicstyle=\linespread{.85}\ttfamily,
                keywordstyle=\color{blue}\ttfamily,
                stringstyle=\color{red}\ttfamily,
                commentstyle=\color{green}\ttfamily,
                morecomment=[l][\color{magenta}]{\#},
}  

\begin{lstlisting}[]
int hash(byte[] key) {
  int i = 0;
  int hash = 0;
  while (i != key.length) {
    hash += key[i++];
    hash += hash << 10;
    hash ^= hash >> 6;
  }
  hash += hash << 3;
  hash ^= hash >> 11;
  hash += hash << 15;
  return hash;
}
\end{lstlisting}


Once word vectors are constructed, similarities between them can be computed using correlation measures (ideally invariant to shift and scale) such as Goodman and Kruskal's $\gamma$ coefficient \cite{GKGamma} (or, alternatively, Kenall's $\tau_b$ \cite{KENDALL01061938}). 
To compute $\gamma$, \emph{concordant} and \emph{discordant} pairs between two vectors must be counted. Given any pairs such as $(x_i, y_i)$ and $(x_{j}, y_{j})$ from two $m$-dimensional vectors $\vec{x}$ and $\vec{y}$ and the value $v=(x_i-x_j)(y_i-y_j)$, these two pairs are concordant if $v>0$ and discordant if $v<0$. If $v=0$, the pair is neither concordant nor discordant. Let $p$ and $q$ be the number of concordant and discordant pairs, then $\gamma$ is given by \newcite[p.~86]{chen2006correlation}:
\begin{equation}\label{eq-gammacorr}
    \gamma = \frac{p -q}{p+q}.
\end{equation}

The loop in the simple algorithm suggested above can be serialized in various ways to meet the requirements of applications for which we build word vectors. For instance, when processing text streams (or, sequential scan of very large corpora), a new vector can be added or an existing one can be removed or updated on the fly. Moreover, since building vectors does not require solving an optimization problem, it can be easily distributed in parallel. For the same reason, this means that, e.g., \citepos{Turney2001} \emph{PMI-IR method}\textemdash which is obsolete, or at best cumbersome to use, in combination with methods based on neural networks\footnote{Or any other method that models word representation learning as an optimization task.}\textemdash now can be used even easier than before. 


As with other statistical models, normalizing the counted frequencies in the obtained word vectors using a `weighting process' usually enhances results. A weighting process usually eliminates uninformative and irrelevant frequencies and boost the impact of informative and discriminatory ones. This can be done through normalizing raw frequencies by the expected and marginal frequencies, and using a threshold for removing uninformative components such as the popular PPMI weighting \cite{Turney2001,Bouma2009} (or, e.g., the less known odds ratio measure).
Since weighting process is carried out over vectors of reduced dimension $m$, the weighting process requires a small computation, too. To facilitate the weighting process in online applications (i.e., when vectors must be updated frequently), one $m$-dimensional vector that holds the sum of coordinates of vectors (or a random subset of them) can reside in memory at all time and the weighting process can be done on demand. When using weighted vectors (which often have rectified Gaussian distribution), e.g. in word semantic similarity tasks, we replace $\gamma$ with a correlation measure such as Pearson's $r$:
\begin{equation}\label{eq-pearson}
    r(\vec{x},\vec{y}) = \frac{\sum_{i=1}^{m}{(x_i - \bar{x})}(y_i-\bar{y}))}{\sqrt{\sum_{i=1}^{m}{(x_i-\bar{x})^2}}\sqrt{\sum_{i=1}^{m}{(y_i-\bar{y})^2}}},
\end{equation}
in which $\bar{x}$ and $\bar{y}$ are the mean of vectors, i.e. e.g., $\bar{x}=\frac{1}{m}\sum_{i=1}^{m}x_i$.


\subsection{Method's Justification}\label{secjustify}

The proposed method can be explained mathematically using the principles of dimensionality reduction using random projections. 

Let $\mathbf{C}_{p \times n}$ denotes the set of $p$ word vectors obtained by counting the co-occurrences of each target word $\vec{w_i}$ ($1\leq i \leq p$) with each context element $c_j$ ($1\leq j \leq n$).
In most applications, $n$ is a very large number in a way that it hinders subsequent processes (and causes the so-called \emph{curse of dimensionality}).
To address this problem, $\mathbf{C}_{p \times n}$ undergoes a set of `transformations' $\mathbf{T}$ to map the high-dimensional space $\mathbf{C}_{p \times n}$ onto a space of reduced dimensionality $\mathbf{W}_{p \times m}$ ($m \ll n$): $
\mathbf{C}_{p \times n} \times \mathbf{T}_{n \times m} = \mathbf{W}_{p \times m}.
$
In our proposed method, $\mathbf{T}_{n \times m}$ is a highly sparse matrix generated randomly, in which $t_{ij}$ elements of $\mathbf{T}$ has the following distribution:
\begin{equation}
t_{ij}= 
\begin{cases} 
0 & \textnormal{with probability } \frac{m-1}{m} \\
1 & \textnormal{with probability } \frac{1}{m}\\
\end{cases}, \label{distrib-elements}
\end{equation}
such that each row vector of $\mathbf{T}$ has exactly \emph{one} component that has value 1, and these non-zero values of $\mathbf{T}$ are independently and identically distributed. It can be verified that the algorithm proposed in the previous section computes the desired word vectors (i.e., $\mathbf{W}$) by (a) de-randomization of $\mathbf{T}$ using a hash function and the modulus operator, and (b) serializing the involved multiplication for computing $\mathbf{C} \times \mathbf{T}$ to a set of addition operations (based on the distributive property of multiplication over addition). 

Novelties of our method is in the way that we design and compute $\mathbf{T}$. Previously, random-projection-based methods\footnote{In general, most dimension reduction methods, e.g., SVD/PCA truncation and so on.} compute and propose a projection $\mathbf{T}$ with the goal of having the \emph{least distortions in pairwise ($\alpha$-normed) distances}\footnote{Or, in general, reconstruction errors computed with respect to a loss function.} when mapping vectors from $\mathbf{C}$ onto $\mathbf{W}$\textemdash e.g., the well-known \citepos{johnson84extensionslipschitz} lemma for $\ell_2$-normed spaces, and the one proposed by \newcite{Indyk2000} for $\ell_1$-normed spaces, and their subsequent refinements and generalizations to $\ell_\alpha$-normed spaces by \newcite{Li2006VSR}. In contrast to the aforementioned projections, we disregard this classic desiderata of preserving distances and the goal of minimum-distortion correspondence. In return, we motivate our proposed random projection directly using the implications that Harris' Distributional Hypothesis bears as stated earlier in the introduction. 

\section{Related Work}\label{relatedwork}
Random projections and hash kernels have been vibrant research areas in theoretical computer science and artificial intelligence thus NLP. These methods have been employed to provide viable solutions for a range of problems that require a notion of (approximate) nearest neighbor search, in particular in information retrieval tasks such as identification of duplicate and near duplicate documents \cite{Manku2007} and string matching \cite{Dalvi2013,citeulike2706665}, semantic labeling \cite{YangLC16}, and cross-media retrieval \cite{Wang2015}, to name but a few. Naturally, these methods also have been applied to the problem of learning word representation, either as a dimensionality reduction method \cite{Bingham2001,kaski1998}, or in the form of an incremental vector space construction technique in which the random projection-based dimensionality reduction process is merged with the process of collecting word co-occurrence frequencies, e.g., \newcite{kanerva00random,qzadehhandschuh2014EMNLP2014} as well as \newcite{vandurme-lall:2010:Short,Geva2011TTP}. 

\citeauthor{kanerva00random} employed \emph{sparse Gaussian random projections} (which preserve pairwise $\ell_2$ distances)  
to build word vectors directly at a reduced dimensionality and showed that their proposed \emph{random indexing} technique yields results comparable to \citepos{DeerwesterDLFH90} \emph{latent semantic analysis technique} which employs singular value decomposition truncation (also a $\ell_2$ preserving dimension reduction method but based on a deterministic algorithm). By the same token, \citeauthor{qzadehhandschuh2014EMNLP2014} extended the idea proposed by \citeauthor{kanerva00random} for the comparison of similarities in $\ell_1$-normed spaces, i.e., for estimating the city-block distance. However, when it comes to comparing semantic similarities between words, these methods fail to compete with the more recent neural-based embedding techniques\textemdash e.g., \citepos{Mikolov2013} word2vec and \citepos{glove2014} GloVe. 

It is known that crude distances between words (as preserved by the aforementioned methods) are not discriminatory enough in semantic similarity assessment tasks. This problem could be alleviated using a weighting process (e.g., PPMI as mentioned earlier) but unfortunately since the aforementioned methods use projections with zero expectation, the sum of components in their resulting vectors is \emph{always} zero; and, consequently, weighting techniques such as PPMI cannot be applied to their resulting models after their construction (simply, due to the problem of division by zero). Note that for these methods, projections with zero expectation is essential for achieving an acceptable performance when computing projected spaces (i.e., to guarantee the sparsity of randomly-initialized projection matrices).\footnote{That is to say, additive smoothing \cite[e.g., see][]{baroni-lenci-Cognitive-2007} is inappropriate due to the computational cost that it imposes.} Compared to these techniques, the method proposed in this paper has a) a better computational complexity (thanks to its sparser projections) and b) yields better results in semantic similarity tasks because of the possibility of applying weighting techniques after the construction of randomly projected spaces. Moreover, the proposed hash-based derandomization eliminates the need for storing random projection matrices used in these methods.  

Last but not least, we must mention representation learning methods such as \cite{ravichandran-pantel-hovy:2005:ACL,vandurme-lall:2010:Short,Geva2011TTP,vandurme-lall:2011:ACL-HLT2011} that are based on (bit-wise) \emph{locality sensitive hashing} (LSH) \cite{Indyk1998:ANN,Charikar:2002:SET}. LSH methods, be it \emph{data-dependent} or \emph{data-independent}, have a strong relationship to random projections (ibid). A good example of a data-independent LSH method designed for preserving distances in $\ell_2$-normed spaces is the \emph{Reservoir Counting} method \cite{vandurme-lall:2011:ACL-HLT2011}. \citepos{vandurme-lall:2011:ACL-HLT2011} method is similar to ours in the sense that it can be used in online settings; however, it differs from our method in that it (like random indexing and \citepos{ravichandran-pantel-hovy:2005:ACL} method) preserves pairwise cosine similarities. Similar to these methods, \newcite{Geva2011TTP,ChappellG15} propose a method for preserving pairwise Hamming distances. Concerning data-dependant LSH, a sophisticated example can be found in \cite{Xu:2015:CNN:2832415.2832440}; these methods, like the well-known neural network-based peers (e.g., GloVe and word2vec) are not comparable to our method in that they demand solving an optimization problem to build word vectors.

\section{Evaluation}\label{sec:evaluation}
We report results from empirical evaluations in both `intrinsic' and `extrinsic' setups. Besides an intrinsic evaluation (i.e., using word vectors obtained by our hash-based method in semantic similarity tests), we report results when these vectors are used as input for training neural networks.  

To ease replication of results reported in this section, unless otherwise stated, we use \citepos{polyglot:2013:ACL-CoNLL} tokenized Wikipedia dumps for training,\footnote{Available for download from \url{https://sites.google.com/site/rmyeid/projects/polyglot}.} and choose context-windows of size 10+10 to collect co-occurrence counts. 

In all the tasks, as baselines, we use word vectors built using the \textsc{word2vec CBOW} algorithm\footnote{Codes obtained from \url{http://word2vec.googlecode.com/svn/trunk/}.} as well as \textsc{GloVe}\footnote{Obtained from \url{https://nlp.stanford.edu/projects/glove/}, v. 1.2.}, for which we set vector dimensionality to 500.\footnote{Put aside its impact on the required time for training, for our input corpus and the targeted tests, we observed that using a dimensionality larger than 500 has an adverse effect on results obtained from both algorithms.} For both models, we stop training after three epochs, and set minimum vocab frequency threshold to 5; the remaining hyperparameters are left to default values. For these vectors, we use cosine as the similarity estimator.






\subsection{Intrinsic Evaluation}\label{eval-intrinsic}

\begin{table*}[t]
\centering

\resizebox{1\linewidth}{!}{
\begin{tabular}{|c|c|l|ccccccccccc|cr|}\cline{4-16}
\multicolumn{1}{c}{}&\multicolumn{2}{c}{}& \multicolumn{11}{|c|}{\textbf{Similarity Tests}} &\multicolumn{2}{c|}{Mean}\\ \cline{2-16}
\multicolumn{1}{c|}{}&\multicolumn{2}{c|}{\textbf{Vector Set}} & \textbf{YP130} & \textbf{RW}   & \textbf{M287} & \textbf{M771} & \textbf{MC30} & \textbf{MEN}  & \textbf{WS353} &   \textbf{WSS} & \textbf{WSR} & \textbf{RG65} &\textbf{SL} &\textbf{A}&\textbf{G}\\ \hline
\multirow{4}{*}{\rotatebox[origin=c]{90}{\small \bf Baseline}}
&\multicolumn{2}{c|}{\textbf{\small Kanerva's RI-Avg}} & 0.14 & 0.06 & 0.17 & 0.16 & 0.47 & 0.25 & 0.28 & 0.4 & 0.16 & 0.46 & 0.11 & 0.242 & 0.203\\
&\multicolumn{2}{c|}{\textbf{\small Classic-{\tiny {\bf Unweighted}}}} & 0.13 & 0.07 & 0.17 & 0.16 & 0.49 & 0.25 & 0.28 & 0.4 & 0.17 & 0.43 & 0.12 & 0.243 & 0.207\\
&\multicolumn{2}{c|}{\textbf{\small Classic-PPMI} }		 &  0.43 & 0.32 & 0.53 & 0.56 & 0.8 & 0.68 & 0.5 & 0.55 & 0.62 & 0.8 & 0.29 & 0.553 & 0.527\\
&\multicolumn{2}{c|}{\small \textbf{W2V-CBOW}} 			 & 0.43		&  0.24		&  0.53	&  	   0.64 &  	  0.72	   &        0.76  &     0.70    &    0.78   &    0.65 & 0.75  &  0.40 &  0.600   & 0.569\\
&\multicolumn{2}{c|}{\small \textbf{GloVe}} 			 &
0.53 &	0.33 &	0.55 &	0.67 &	0.76 &	0.74 &	0.67 &	0.76 &	0.42 &	0.81 &	0.37 &
0.600   & 0.576
\\\hline\hline
\multirow{8}{*}{\rotatebox[origin=c]{90}{ \bf Our-Method}}&\multirow{4}{*}{\rotatebox[origin=c]{90}{ \small  Unweighted}}
 &{\bf \small Dimension=500}&0.24 &	0.09 &	0.14 &	0.48 &	0.75 &	0.59 &	0.48 &	0.59 &	0.41 &	0.77 &	0.29 &	0.439&	0.365\\
&&{\bf \small Dimension=1000}&0.30 &	0.21 &	0.39 &	0.40 &	0.75 &	0.58 &	0.49 &	0.53 &	0.49 &	0.76 &	0.29 &	 0.471 & 0.439\\
&&{\bf \small Dimension=2000}&0.29 &	0.21 &	0.39 &	0.40 &	0.74 &	0.58 &	0.49 &	0.52 &	0.48 &	0.78 &	0.29 &	 0.469 & 0.437\\
&&{\bf \small Dimension=4000}&0.29 &	0.21 &	0.39 &	0.40 &	0.73 &	0.58 &	0.48 &	0.51 &	0.48 &	0.79 &	0.29 &	 0.469 & 0.437
\\\cline{2-16}
&\multirow{4}{*}{\rotatebox[origin=c]{90}{  PPMI}}
 &{\bf \small Dimension=500}&0.35 &	0.31 &	0.52 &	0.53 &	0.80 &	0.69 &	0.66 &	0.71 &	0.66 &	0.75 &	0.31 &		0.572 &	0.543\\
&&{\bf \small Dimension=1000}&0.40 &	0.33 &	0.58 &	0.56 &	0.81 &	0.72 &	0.68 &	0.75 &	0.68 &	0.80 &	0.32 &	 0.604 & 0.574\\
&&{\bf \small Dimension=2000}&0.42 &	0.32 &	0.61 &	0.57 &	0.81 &	0.73 &	0.69 &	0.76 &	0.68 &	0.79 &	0.32 &	 0.609 & 0.580\\
&&{\bf \small Dimension=4000}&0.43 &	0.31 &	0.62 &	0.58 &	0.83 &	0.73 &	0.70 &	0.77 &	0.69 &	0.83 &	0.32 &	 0.618 & 0.587\\
\hline
\end{tabular}
 }
\caption{Performance of our method in semantic similarity and relatedness tasks and its comparison to baselines. The last two columns report arithmetic and geometric mean of the observed performances (i.e., the harmonic mean of $\rho$ and $r$) obtained across tasks. 
For the hash-based method, we report results for both unweighted and PPMI-weighted vectors of various dimensionality. We use Goodman and Kruskal's $\gamma$ and Pearson's $r$ for estimating similarities for unweighted and PPMI weighted vectors build by our method, respectively. For baselines, we use cosine as the similarity estimator. 
}
\label{test-1}
\end{table*}

\subsubsection{Word Similarity Datasets} \label{res-sim-test}

We evaluate our method over a number of word similarity tests. Each test encompasses a set of word pairs, associated with similarity/relatedness ratings obtained from human annotators. For each test, evaluation takes the form of calculating \emph{the harmonic mean} of Pearson $r$ and Spearman's $\rho$ correlations between the list of word pairs sorted by the human-induced scores and the one sorted by the scores assigned to word pairs by using similarities between word vectors. We run our experiments over \textsc{Ws353} \cite{Finkelstein2001}, \textsc{Wss} and \textsc{Wsr} by \cite{Agirre2009}, the classic tests of \textsc{Mc30} \cite{MC30} and \textsc{Rg65} \cite{RG1965}, the Stanford Rare Word dataset \textsc{RW} \cite{rw2013}, \textsc{M287} by \cite{Radinsky2011}, \textsc{M771} \cite{Halawi2012}, \textsc{SL} \cite{Hill2015}, \textsc{YP130} verb relatedness \cite{Yang06}, and \textsc{Men} \cite{Bruni2014}. 

In addition to results obtained from the CBOW and GloVe models, we report also the performance of a classic un-weighted and PPMI-weighted high-dimensional model (the so-called \emph{count} model) as baselines. Moreover, to validate our earlier statement about the distance preservation property of sparse Gaussian random projections, we also build vectors using \citeauthor{kanerva00random}'s random indexing method (we use projection (index) vectors of dimension 2000 in which which four components are set to +1 and another four to -1).

For our method, we build word vectors of dimension $m=$ 500, 1000, 2000, and 4000, and report results obtained from both the unweighted vectors\textemdash with $\gamma$ (Eq.~\ref{eq-gammacorr}) used as the similarity estimator\textemdash as well as the PPMI-weighted vectors\textemdash with Pearson's $r$ (Eq.~\ref{eq-pearson}) as the similarity estimator. Table~\ref{test-1} summarizes the observed results; to ease comparison, we report the arithmetic and geometric mean of performances across tasks, too.

\begin{table*}
\centering
\resizebox{.95\textwidth}{!}{
\begin{tabular}{|l|lllllllllll|}
\hline
&YP           & RW     & M287 & M771  & MC30  & MEN  & WS353 & WSS   & WSR  & RG65  & SL         \\\hline
\textbf{\textsc{Hash} - CBOW}   & 1.06   & 3.41 & 7.11  & -2.77 & 2.36 & 2.42  & 5.83  & 2.05 & 12.51 & 4.61  & -7.12 \\
\textbf{\textsc{Hash} - GloVe} & -17.35 & 3.72 & 3.56  & -6.21 & 8.92 & -0.7  & 11.23 & 5.71 & 14.35 & 0.89  & -4.81 \\
\textbf{CBOW - GloVe}  & -18.4  & 7.12 & -3.17 & -3.44 & 6.56 & -3.13 & 5.4   & 3.66 & 1.83  & -3.72 & 2.35 \\\hline
\end{tabular}
}
\caption{Differences in Spearman's $\rho$s obtained by different methods; \textsc{Hash} denotes our method.}\label{tab-diff-rho}
\end{table*}

Regarding the unweighted and PPMI-weighted count models, the effective dimension of vectors soared to a little more than $5.4$ million. As expected, we witnessed that the obtained averaged results  (over 5 independent runs) from the random indexing method are almost identical to those observed and reported for the unweighted high-dimensional classic model. However, the random indexing has the advantage of delivering, more or less, the same results with vectors built at a considerably lower dimensionality.

Concerning our model, disregarding $m$, our unweighted and PPMI-weighted models outperform their counterpart classic high-dimensional models, as well as the random indexing method. Moreover, for large values of $m$ (e.g., $m=1000$), our hash-based PPMI-weighted vectors (for most tasks) are a match for the word2vec's CBOW and GloVe models, particularly when considering the statistical significance of the difference between the observed results from these models, as it is described by \newcite[][see Table 2]{rastogi-vandurme-arora:2015:NAACL-HLT} and their proposed \emph{Minimum Required Difference for Significance} (MRDS) measure. Here, we use MRDS to filter out insignificant comparative gains between our method and the baselines. Table~\ref{tab-diff-rho} lists the difference between the \emph{Spearman $\rho$ correlations} obtained by the vector sets built using the word2vec CBOW method, GloVe, and our method. Based on MRDS values in  \newcite{rastogi-vandurme-arora:2015:NAACL-HLT}, while our method lags behind \textsc{CBOW} and \textsc{GloVe} in \citepos{Hill2015} SL, it certainly performs better than its neural counterparts in the \textsc{RW} and \textsc{Wsr} tests. Similarly, MRDS suggests that in the \textsc{Men} test, the \textsc{CBOW} clearly outperforms all the other methods, whereas the difference in the obtained $\rho$s by our method and GloVe is insignificant. By the same token, \textsc{GloVe} and \textsc{CBOW} both perform equally well over most of the tests except \textsc{MEN} and \textsc{RW}, for which CBOW and GloVe perform better than the other, respectively.\footnote{We are missing MRDS values for \textsc{YP} and \textsc{M771}.}

\subsubsection{Enhancing Results in Similarity Tasks}

The performance of our proposed vector representation learning method in semantic/relatedness similarly tasks can be improved in several ways. The easiest method is perhaps to use models of large dimension (as implied by results reported in the previous section), or/and to enlarge the size of input corpora.

As reported in Table~\ref{test-1}, we observe that increasing the dimension of our hash-based vectors (coupled with PPMI weighting) enhances results in similarity/relatedness tests. Obviously, increasing the dimension of our vector  does not affect the computational complexity (nor the required training time) for building a model; however, it increases the cost of similarity computation. Figure~\ref{dimensionAvgPerformance} plots changes in the method's performance (i.e., the arithmetic average of performances across tasks in our experiments) when the dimensionality of our model is increased\textemdash the evaluation setup (i.e., the input corpus and context window size) remains the same as those reported earlier. As shown, up to $m=10,000$, the performance mostly improves. We admit that increasing dimensionality is not desirable due to the cost that it imposes in the subsequent process (e.g., estimating pairwise similarities). However, we notice that these models of higher dimensionality (e.g., $m=2000$) can be compressed (e.g., to $m=250$) using matrix factorization techniques (e.g., PCA) without hurting task-based performances. Since our models are already of low-rank, factorization (e.g., computing principal components) can be done considerably fast.  


To show the effect of enlarging the size of input corpus on the method's performance, in addition to the Wikipedia corpus (1.8 billion) that we used initially, we fed another 4 billion tokens of web crawled text data \cite{SchaeferBildhauer2012} to our models. As expected, disregarding the model's dimensionality (here 500, and 2000), using larger input corpora enhances the method's performance in similarity/relatedness tasks, as shown in Figure~\ref{corpusSizeFig}. Evidently, enlarging the input training corpus also can enhance results obtained by other baseline methods. However, training and updating our model with new text can be resumed and done much easier compared to these methods.

\input{fig-perform-hyper-param.tex}

Additionally, like many word representation techniques, our method can be tuned for each task too, e.g., by choosing an appropriate context window size, and in general, by making (let's call it) linguistically more informed decisions. For instance, we could boost the result for the \textsc{Men} test to 0.77 using larger context window (i.e., 13+13 instead of 10+10) and filtering context words that appear in a stop-word list. Similarly, we could enhance the obtained performance for the \textsc{YP130} verb relatedness from 0.40 to 0.69 (for $m=500$) through filtering context words using dependency parses (context elements were limited to those words with direct syntactic relationships to the targeted verbs). Apart from these, our method allows for heterogeneous context types: For example, in addition to context words, we use document-level co-occurrences to boost performance in the WS353 test from 0.66 to 0.71. For each $w$ appeared in document $d_i$, we pass to the hash function $d_i$'s identifier in our input and continue updating $\vec{w}$ as instructed in the algorithm. Note that there is no limit on the type of context elements that can be fed to our method as long as these elements can be converted to byte sequences and subsequently to hash codes. We suggest that similar improvements can be gained using sub-word features, e.g., character n-grams extracted from context words. In comparison, this is not that straightforward with neural representation learning techniques; these sorts of changes usually demand a change in their objective/loss function (which defines their underlying optimization goal).

 \emph{Retrofitting}\textemdash i.e., to refine word vectors using lexical relations available in lexical resources \cite{FaruquiDJDHS15}\textemdash is another simple yet effective methodology for improving results obtained in similarity/relatedness tests. We adapt the idea in its simplest form and treat lexical resources like any other text files; but, instead of scanning these resources using a sliding context window, we make sure that we encode co-occurrence information about all related items in each entry of an input lexical resource. For example, given a synset $S$ of words in WordNet (i.e, $S=\{w_1 \ldots w_n\}$), for each $w_i \in S$, we consider all the remaining words in $S$ as co-occurring context elements and respectively update $\vec{w_i}$ as described earlier in Algorithm~\ref{algorithm1}.  To show the impact of this `retrofitting' method, we take vectors of dimension $m=500$ from our earlier experiment, and update them by reading WordNet \cite{Miller1995} and PPDB \cite{ganitkevitch2013ppdb} (i.e., in addition to the Wikipedia corpus, WordNet and PPDB are also fed to our algorithm). As Table~\ref{retrofit} reports, feeding these resources to our model yields a considerably larger performance gain, e.g., compared to CBOW. While CBOW and GloVe requires an inefficient re-training, our method simply updates the counts by scanning the new input.

\begin{table*}
\centering
\resizebox{1\linewidth}{!}{
\begin{tabular}{|c|ccccccccccc|cr|}\cline{2-14}
\multicolumn{1}{c}{}& \multicolumn{11}{|c|}{\textbf{Similarity Tests}} &\multicolumn{2}{c|}{Mean}\\ \cline{1-14}
\textbf{Vector Set} & \textbf{YP130} & \textbf{RW}   & \textbf{M287} & \textbf{M771} & \textbf{MC30} & \textbf{MEN}  & \textbf{WS353} &   \textbf{WSS} & \textbf{WSR} & \textbf{RG65} &\textbf{SL} &\textbf{A}&\textbf{G}\\ \hline
\textbf{\small W2V-CBOW} &0.46 &	0.24 &	0.52 &	0.63 &	0.74 &	0.75 &	0.70 &	0.78 &	0.66 &	0.77 &	0.38&   0.602  &  0.571  \\
\textbf{\small GloVe} & 0.69 &	0.35 &	0.58 &	0.66 &	0.80 &	0.73 &	0.64 &	0.74 &	0.60 &	0.81 &	0.41 & 0.637 & 0.619\\
{\bf Hash}		 & 0.76 &	0.31 &	0.51 &	0.59 &	0.82 &	0.70 &	0.60 &	0.73 &	0.52 &	0.78 &	0.53 &		0.625 &	0.605\\
\hline
\end{tabular}
 }
\caption{Results from retrofitting: All the methods build vectors of dimensionality 500, and they are trained on the Wikipedia corpus as well as WordNet and PPDB. Comparing these results to those reported in Table~\ref{test-1} shows a considerable improvement of our method's performance. }
\label{retrofit}
\end{table*}

\subsubsection{Linear substructures}
The performance of word vector learning method in the word analogy task \cite{mikoAna} is often consider as the presence of `meaningful substructure' in the learned spaces. E.g., it is expected that a model encodes the analogy `king is to queen as man is to woman' such that 
$\vec{\mathit{king}}$ $-$ $\vec{\mathit{queen}}$ = $\vec{\mathit{man}}$ $-$ $\vec{\mathit{woman}}$. To our disappointment, vectors obtained from our hash-based method (at least out-of-the-box) do not perform  as well as vectors obtained using CBOW and GloVe. 

Namely, GloVe yields a `semantic' accuracy of $33.70\%$ (403/1196) and a `syntactic' accuracy of $55.10\%$ (5350/9709); for CBOW, semantic accuracy is $24.24\%$ (290/1196) and syntactic accuracy is $45.35\%$ (4404/9709). However, our method (PPMI-weighted vectors of dimension 500, Pearson's $r$ as similarity estimator)\footnote{The same sets of vectors used \S~\ref{res-sim-test}.} gives a semantic accuracy of $13.93\%$ (344/2469) and a syntactic accuracy of $35.44\%$ (3582/10106).\footnote{Note that when building vectors, CBOW and GloVe discarded words with a frequency lower than 5, hence they cover less words from of the analogy task vocab.} Perhaps methodologies proposed by \newcite{Levy2015improving} could improve the performance of our method, which we leave for for future investigations.

\subsubsection{Comparing training time}
Concerning the training time, our method is noticeably faster than all the other baseline methods used in this paper. Particularly, our method generates vectors at least twice as fast as the C implementation of the \emph{CBOW} algorithm. Namely, our multithread implementation\footnote{Based on Java 1.8 \texttt{AtomicLongArray} and \texttt{ConcurrentHashMap} data structures and parallel streams for \texttt{java.nio.file} interface.} builds vectors of dimension 500 for all the words (around $13$ million) appeared in our Wikipedia training corpus in slightly less than 67 minutes, whereas the CBOW training time using the same number of processing cores for only 1 epoch of training is 124 minutes (CBOW and GloVe build vectors only for 1,778,575 words). Given the way processes are factorized in the C implementation of GloVe, we cannot provide a similar comparison for it; least to say that in the worst case, our method will take slightly longer\footnote{Given the required additional operations for the hash code generation in our method.} than the its required \texttt{cooccur} preprocessing (i.e., the step to constructs word-word co-occurrence statistics from a corpus). Compared to \texttt{cooccur} process, our method requires less memory since we do not keep track of the exact word-word counts (this is also true when comparing our method to the classic high-dimensional model). Our method is much faster than random indexing, too, thanks to the fewer mathematical operations that it requires (i.e., one addition operation vs minimum two operations for random indexing). 

Concerning the weighting process, while it takes hours to build PPMI-weighted vectors for only a fraction of the vectors in the high-dimensional count model (e.g., the vocabulary of size $~6000$ used in \S~\ref{res-sim-test}), for the same vocabulary it only takes a few second to convert raw frequencies to PPMI weights.

\subsection{Extrinsic Evaluation}
Word vector representations are an important ingredient of upstream neural-network-based NLP systems. Put simply, they act as input feature structures for the underlying neural networks. In this context, to compare vectors obtained by our method with more established choices such as word2vec and GloVe, we use the trained vectors from \S~\ref{res-sim-test} as input for neural networks applied to a sentence-level classification task, i.e., the binary sentiment classification over the {\it Large Movie Review Dataset} \cite{maas-EtAl:2011:ACL-HLT2011} using a simple convolutional neural network (CNN) model and a recurrent neural network (RNN) model. Per suggestion in \cite{kim:2014:EMNLP2014}, in all the experiments, we keep the pre-trained word vectors static and learn only the other parameters of the CNN and RNN models.

Our CNN model is an implementation of the \emph{CNN-static} model proposed in \cite{kim:2014:EMNLP2014}; the RNN model simply replaces the CNN layer with a LSTM one. Both models were previously tuned on a development set using \citepos{mikoAna} pre-trained vectors of dimensionality 300 (trained over a Google news corpus). For our experiments, we simply replace these vectors with the ones that we trained earlier over the Wikipedia corpus and perform \emph{no} parameter tuning. Each model is trained 4 times for two epochs using each of the vector sets and the average of the obtained f1-scores are reported as the overall performance. We make sure that all the vector sets (i.e., PPMI-weighted vectors from our method of dimension 500 and 1000, CBOW, and GloVe) cover the same vocabulary (i.e., 66,626 out of 89,527 words in the dataset's vocabulary).

Table~\ref{nn-based-result} provides a summary. Our method clearly outperforms GloVe in all the settings with a large margin. The poor performance of GloVe is particularly surprising given that it showed a relatively better performance than the other two methods in the intrinsic evaluations (specially the analogy task). Concerning the CBOW model, vectors built by our method could outperform the CBOW baseline when using the CNN model, provided that we use vectors of the larger dimenionality of 1000 (which consequently comes at the expense of a longer training time). Regarding the RNN model, the CBOW model outperforms our method particularly for vectors of low dimesnionality. As we increase the size of vectors obtained by our model, the performance gap between the two methods decreases.

\begin{table}
\centering
\resizebox{.95\columnwidth}{!}{
\begin{tabular}{l|ll|ll|}
\cline{2-5}
                                  & \multicolumn{2}{c|}{\bf CNN}                             & \multicolumn{2}{c|}{\bf RNN}                             \\ \cline{2-5} 
                                  & \multicolumn{1}{c|}{avg} & \multicolumn{1}{c|}{Time} & \multicolumn{1}{c|}{avg} & \multicolumn{1}{c|}{Time} \\ \hline
\multicolumn{1}{|l|}{\textbf{GloVe}}       & 83.92                    & 27,19                     & 73.60                    & 59,53                     \\ 
\multicolumn{1}{|l|}{\bf CBOW}        & 85.14                    & 27,39                     & \textbf{80.27}           & 66,20                     \\ 
\multicolumn{1}{|l|}{\bf Hash m=500}  & 84.66                    & 26,8                      & 77.73                    & 59,29                     \\
\multicolumn{1}{|l|}{\textbf{Hash m=1000}} & \textbf{86.54}           & 40,22                     & 79.86                    & 72,83                     \\ \hline
\end{tabular}}
\caption{Comparison of our method against Glove and CBOW with respect to results obtained from the sentence-level classification task. Avg. denotes the arithmetic mean of the obtained f-scores across 4 independent runs. The column `Time' shows the average training time in minutes. }
\label{nn-based-result}
\end{table}

\section{Conclusion}\label{conclusion}
Recently, \emph{prediction-based} representation learning methods attract substantial attention based on the argue that these models perform better across a range of tasks
support. In this work we show that comparable results can be attained using an algorithm that avoids solving an optimization problem required by prediction-based representation learning methods. Our presented method is based on a new random projection that goes beyond the usual goal of pairwise distance preservation. It is fast and flexible and requires a small amount of computational resources to learn/build a model. Yet, as shown empirically, it shows performance competitive to prediction-based methods in both intrinsic and extrinsic evaluation setups.

\input{ijca2017.bbl}
\balance

\end{document}

%% file: fig-perform-hyper-param.tex
\pgfplotstableread{parameters/dim-avg-performanc.txt}\dimPertableToefl
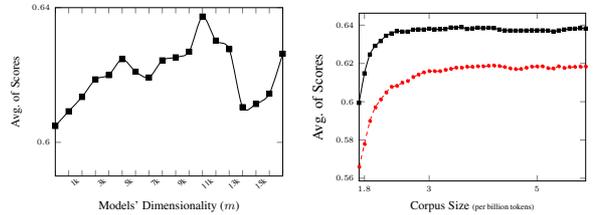
\begin{figure}
	\centering
\subfloat[Variable Dimensionality]{
	\resizebox{.49\columnwidth}{!}{
		\begin{tikzpicture}
		\begin{axis}[
		every tick label/.append style={font=\tiny},
		legend style={at={(1.0,0.2)}},
		ylabel= {\small Avg. of Scores},
		xlabel= {\small Models' Dimensionality ($m$)},
		height=6cm,width=7.5cm,
		xtick=data,
		scaled x ticks=false,
		xticklabels ={,, 1k, , 3k,,5k,,7k,,9k,,11k,,13k,,15k,,17k},
		x tick label style={rotate=40,anchor=east},
		axis on top,
		xmin=1000,
		ytick={0.60,0.64},
		xmax = 18000,
		ymax=0.64,
		ymin=0.59,
	]
	\addplot[name path=PAve,black, mark size=1.9, mark=square*,   black, smooth] table[x=dim,y=avgperform2] from \dimPertableToefl;
	\end{axis}
	\end{tikzpicture}
	}
	\label{dimensionAvgPerformance}
	}
\pgfplotstableread{parameters/corpus-size.txt}\dimPertableSize
\pgfplotstableread{parameters/corpus-size2.txt}\dimPertableSizeFive
\subfloat[Variable Corpus Size]{
	\resizebox{.49\columnwidth}{!}{
		\begin{tikzpicture}
		\begin{axis}[
		every tick label/.append style={font=\tiny},
		legend style={at={(1.0,0.2)}},
		ylabel= {Avg. of Scores},
		xlabel= {\small Corpus Size {\tiny(per billion tokens)}},
		height=6cm,width=7.5cm,
		xmin=1.7,
		xmax=5.9,
		xtick={1.8,3,5},
		axis on top,
	]
	\addplot[mark size=1.2, mark=square*,name path=PAve,smooth, black] table[x=size,y=avg1] from \dimPertableSize;
	\addplot[smooth, dashed, mark size=1.2, mark=*,name path=PAve,red] table[x=size,y=avg1] from \dimPertableSizeFive;
	\end{axis}
	\end{tikzpicture}
	\label{corpusSizeFig}
	}
	
	}
\caption{Changes of the averaged performances across tasks when increasing dimensionality of models (\ref{dimensionAvgPerformance}), and increasing the size of input corpus (\ref{corpusSizeFig}). In Figure \ref{corpusSizeFig}, the black and red lines plot averaged performances for models of dimensionality $m=$500 and 2000, respectively.
}\label{performance-dimension}
\end{figure}